%% file: main_mlhc_final.tex
\documentclass[pmlr]{jmlr}
\RequirePackage{graphicx}
\usepackage{subcaption}
\usepackage{multirow}
 \usepackage{booktabs}
 \usepackage{rotating}
 \usepackage{enumitem}
 \usepackage{cleveref} 
\usepackage{longtable}
\usepackage{bbm}
\usepackage{lipsum}
\usepackage{pdflscape} 
\input{math_commands.tex}

\makeatletter
\def\set@curr@file#1{\def\@curr@file{#1}} 
\makeatother
\usepackage[load-configurations=version-1]{siunitx} 

\theorembodyfont{\upshape}
\theoremheaderfont{\scshape}
\theorempostheader{:}
\theoremsep{\newline}

 \jmlrproceedings{PMLR}{Proceedings of Machine Learning Research}
\jmlrvolume{340}
\jmlryear{2026}
\jmlrworkshop{Machine Learning for Healthcare}

\title[Automated Bowel Obstruction Detection and Localization]{Bowel Obstruction Detection and Localization on Abdominal CT with Deep Learning}

\author{%
\Name{Moritz Vandenhirtz}$^{1,*}$ \Email{moritz.vandenhirtz@inf.ethz.ch}\\[0.1em]
\Name{Andrea Agostini}$^{1,*}$ \Email{andrea.agostini@inf.ethz.ch}\\[0.1em]
\Name{Dana Belde}$^{2,3}$ \Email{dana.belde@ksb.ch}\\[0.1em]
\Name{Mélanie Roschewitz}$^{1}$ \Email{melanie.roschewitz@ai.ethz.ch}\\[0.1em]
\Name{Ismaiel Chikh Bakri}$^{2}$ \Email{ismaiel.chikhbakri@ksb.ch}\\[0.1em]
\Name{Tilo Niemann}$^{2}$ \Email{tilo.niemann@ksb.ch}\\[0.1em]
\Name{André Euler}$^{2}$ \Email{andre.euler@ksb.ch}\\[0.1em]
\Name{Julia E. Vogt}$^{1}$ \Email{julia.vogt@inf.ethz.ch}\\[0.6em]
\addr 
$^{1}$ Department of Computer Science, ETH Zurich, Zurich, Switzerland\\
$^{2}$ Department of Radiology, Kantonsspital Baden, affiliated Hospital for Research and Teaching of the Faculty of Medicine of the University of Zurich, Baden, Switzerland\\
$^{3}$ Department of Forensic Medicine Zurich, University of Zurich, Zurich, Switzerland\\
$^{*}$ Equal contribution
}

\begin{document}

\maketitle

\vspace{-0.2cm}
\begin{abstract}
Bowel obstruction is a common and potentially life-threatening gastrointestinal condition. In the face of rising diagnostic workloads, the automated diagnosis of bowel obstruction on CT scans supports radiologists by accelerating detection and improving patient outcomes. In this work, we propose a deep learning framework with a multi-task objective that jointly detects bowel obstruction and localizes its transition zone. Additionally, we extend the method with an inherently interpretable classification method that locates the suspected transition point within a slice. It does so by learning a probabilistic selection mask that faithfully bases the classifier's prediction solely on a small image region. The proposed method is evaluated on an internal dataset comprising 1,427 abdominal CTs. Here, the model achieves an obstruction detection test accuracy of 93\% and a Hit@10 transition zone localization of 95\%. As the first method to reliably localize the transition zone, this marks a significant step towards the automated identification of this critical clinical landmark.
\end{abstract}
\section{Introduction}
Bowel obstruction is a partial or complete blockage of the small or large intestine that prevents the normal passage of food, fluids, gas, and digestive waste.
This critical gastrointestinal disease causes approximately 260,000 yearly hospital admissions in the US alone~\citep{peery2019burden,peery2022burden,peery2025burden}. The clinical management of bowel obstruction requires rapid and accurate diagnosis, as delays or diagnostic errors contribute directly to the substantial mortality rates associated with the condition~\citep{peery2025burden}. Established guidelines recommend the use of computed tomography (CT) to diagnose suspected bowel obstruction~\citep{ten2018bologna,chang2020acr} with a diagnostic accuracy exceeding 90\%~\citep{megibow1991bowel,frager1994ct}. 

\bigskip\noindent
The dramatic surge in the volume of images that radiologists are required to analyze~\citep{lantsman2022trend,maxwell2021increasing,peng2022radiologist} poses a significant challenge to radiologists, who must meticulously scrutinize and interpret an overwhelming number of images within a limited timeframe. 
Moreover, studies have unveiled a disconcerting correlation between the time allocated to radiologists per image and the accuracy of their diagnostic predictions~\citep{brady2017error, hanna2018effect}. 
Deep learning-based decision-support tools offer the potential to significantly improve patient outcomes by triaging urgent cases and reducing time-to-diagnosis for bowel obstructions. Despite this promise, only a handful of studies investigate the automated detection of bowel obstruction with CT, highlighting a critical gap in the current literature~\citep{vanderbecq2024deep, murphy2024towards}.

\bigskip\noindent
Beyond detecting the presence of a bowel obstruction, reliably localizing its transition zone is critical for clinical intervention~\citep{zins2020adhesive}. The transition zone is the site of the obstruction, whose localization helps identify the cause, determine treatment, as well as plan a potential surgical procedure. 
However, locating this point within a high-resolution CT volume is a time-intensive task that imposes a significant cognitive load on radiologists.
Deep learning-based tools offer a solution by narrowing the search space of possible transition zone locations. By identifying a small subset of candidate locations, such a tool can greatly reduce the number of images that radiologists must manually examine.

\bigskip\noindent
In this work, we introduce a joint framework designed to detect bowel obstructions and identify the transition zone region for positive patients. To the best of our knowledge, this marks the first automated method that jointly tackles these two crucial tasks. To this end, a slice-level classifier with two classification heads is utilized, one for identifying patient-level bowel obstruction and the other for localizing slice-level transition zones. Additionally, we propose an extension based on P2P~\citep{vandenhirtz2025pixels}, a recent advance in interpretable image classification that enables the model to faithfully visualize the region used to make its classification, thereby precisely locating the suspected transition point within a slice. As such, the proposed method performs a multi-grained analysis by detecting bowel obstruction, identifying slices likely to contain the transition zone, and localizing the suspected region within the slices.

\bigskip\noindent
\textbf{Our main contributions}\footnote{The code is available here: \url{https://github.com/agostini335/bo-detection-localization}.} are as follows:
\begin{enumerate}[label=(\emph{\roman*})]
    \item We introduce a novel method that detects bowel obstruction and jointly localizes slices containing the transition zone.
    \item We incorporate an interpretable image classifier that faithfully localizes the specific subregions that drive the model's predictions.
    As such, the proposed method acts as a hierarchical system that (a) detects bowel obstruction on a patient level, (b) finds a set of candidate slices likely to contain the transition zone, and (c) localizes the specific subregions associated with the transition zone.
    \item We provide a rigorous empirical assessment of the proposed method on a clinical dataset, demonstrating the effectiveness of each component in our hierarchical system.
\end{enumerate}

\subsection*{Generalizable Insights about Machine Learning in the Context of Healthcare}
This work demonstrates the utility of medical AI systems that reflect the radiological workflow rather than optimizing for a single task. By designing a hierarchical framework, the proposed method moves beyond a simple binary classification problem to a joint architecture that detects patient-level obstruction while simultaneously localizing slice-level transition zones, thereby supporting radiologists throughout the whole diagnostic process. Additionally, our adaptation of the P2P method to the medical domain illustrates a viable path towards faithful localization of important input regions, such as the transition zone, in a clinical setting.
In the empirical assessment, we showcase the value of a fine-grained evaluation procedure that analyzes each level of the method's diagnostic hierarchy individually, thereby precisely delineating the clinical capabilities of the proposed method.
Finally, by showcasing that bowel obstruction detection and transition zone localization is a solvable task, we hope to inspire more machine learning research into this domain, as it remains vastly underexplored in the current literature.


\section{Related Work}

\subsection*{Deep Learning–Based Approaches to CT Imaging}
Deep Learning methods have been widely applied to CT imaging for tasks including classification and segmentation. Convolutional neural networks (CNNs), particularly Residual Networks (ResNets), introduced by \cite{he2016deep}, are commonly used due to their ability to capture local spatial patterns and learn increasingly complex representations, while maintaining stable optimization through skip connections.
ResNet-based architectures have demonstrated strong performance across a range of CT applications, including lung cancer prediction and COVID-19 diagnosis \citep{kumar2024unified, zhou2023covid}. More recently, transformer-based architectures have been introduced to medical imaging. Vision Transformers (ViTs), proposed by \cite{dosovitskiy2020image}, model images as sequences of patches processed via self-attention, enabling the capture of long-range dependencies and global context. Such models have shown promising results in learning richer representations for CT images \citep{marcos2024pure}.

\subsection*{Deep Learning for Bowel Obstruction}
For a long time, radiography, rather than CT, has been the preferred imaging method for detecting bowel obstruction~\citep{maglinte1996reliability} as it is inexpensive and widely available. Here, leveraging transfer learning with pretrained model such as Inception~\citep{szegedy2016rethinking} has emerged as a robust approach for the automated detection of small bowel obstruction~\citep{cheng2018detection,cheng2019refining,kim2021artificial}. However, current established guidelines recommend the use of computed tomography (CT) to diagnose suspected bowel obstruction~\citep{ten2018bologna,chang2020acr} due to improved obstruction detection~\citep{ahmad2025obstruction} and, most importantly, improved localization of the transition zone~\citep{paulson2015review}.

\bigskip\noindent
While the potential for automated bowel obstruction detection on CT scans is significant, existing literature remains limited. \citet{vanderbecq2024deep} demonstrate that convolutional neural networks can effectively identify bowel obstructions on abdominal CT volumes, offering strong evidence for the feasibility of automated diagnosis. Targetting specific subgroups, \citet{oh2023deep} focus on the detection of high-grade small bowel obstruction and \citet{chang2026interpretable} targets strangulated small bowel obstruction in pediatric patients. Instead of direct detection, \citet{murphy2024towards} segment the gastrointestinal tract, whereby the predicted diameter and longitude can be used to diagnose an obstruction. Similarly, \citet{park2025deep} segment the bowel and show that the manually extracted large bowel length can be indicative of bowel obstruction. Finally, to the best of our knowledge, \citet{vanderbecq2022adhesion} is the only work to attempt the automated localization of the transition zone. Notably, their method is only applied to positive patients, thus, unable to detect bowel obstruction on a patient-level. While their findings are encouraging as a proof of concept, the high rate of false positives suggests significant room for improvement; specifically, an average of 17 out of 125 patches per patient were erroneously classified as containing the transition zone.



\section{Methods}
\label{sec:method}
To align with the standard radiological workflow for suspected bowel obstruction, which relies on a systematic, slice-based analysis of CT volumes, the proposed method introduces a unified framework for clinical decision support. By jointly performing patient-level obstruction detection and slice-level transition zone identification, the model operates natively at the slice level, mirroring clinical practice. In \Cref{subsec:classifier}, the proposed method for detecting the presence of bowel obstruction and the localization of the slices that contain the transition zone, is introduced. Subsequently, \Cref{subsec:p2p_classifier} extends this approach to the localization of the transition point within a given slice by adapting P2P~\citep{vandenhirtz2025pixels}. 

\begin{figure}
    \centering
    \includegraphics[width=1\textwidth]{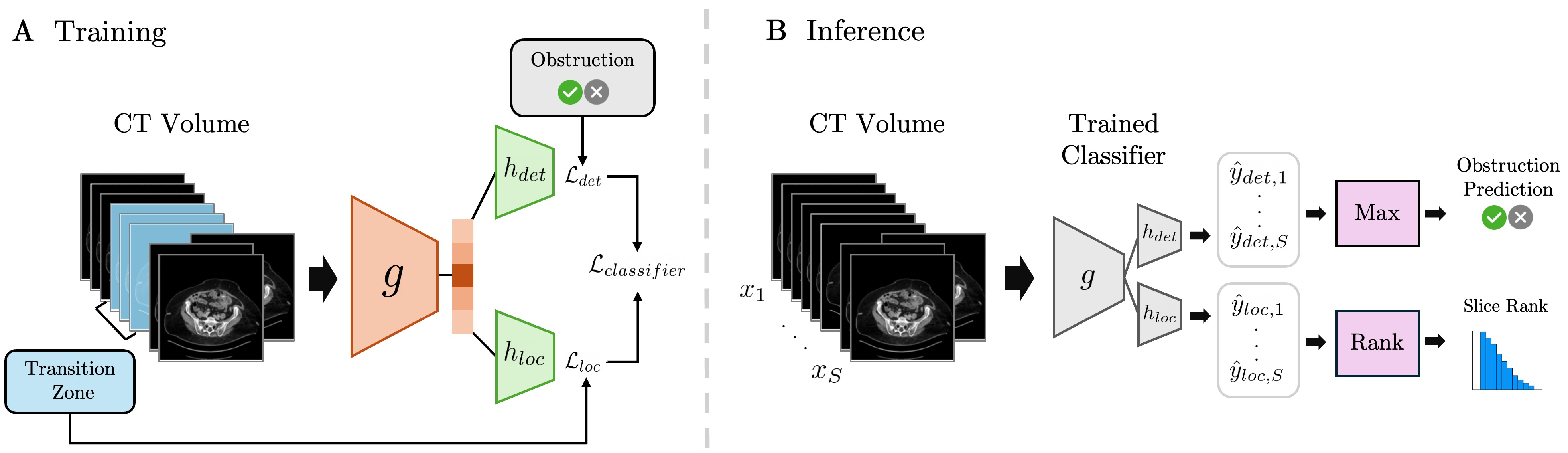}
    \caption{
    \looseness=-1
    Schematic overview of the proposed method. (A) During training, a slice is processed through backbone $g$ and task heads $h_{det}, h_{loc}$. The loss consists of patient-level obstruction detection and slice-level transition zone localization, applied only to patients with obstruction. (B) At inference, all slices are processed individually. Obstruction is determined via max-aggregation of all $\hat{y}_{det}$, while transition zones are identified by ranking all $\hat{y}_{loc}$.}
    \label{fig:classifier}
\end{figure}

\subsection{Obstruction Detection and Transition Zone Localization}
\label{subsec:classifier}
This section outlines the methodology for training a model $f$ that jointly detects bowel obstruction and localizes the transition zone. 
This objective presents a unique challenge, as the operational requirements for the two tasks diverge between the patient-level and slice-level scales.
To resolve this, the proposed model is designed to natively operate on a slice basis, thereby enabling it to leverage the slice-level transition-zone annotations. To achieve patient-level bowel obstruction detection, the inference procedure subsequently combines the slice-level features into a volumetric prediction. A schematic overview is provided in \Cref{fig:classifier}.
The following paragraphs detail the two stages of this approach.

\begin{figure}
    \centering
    \includegraphics[width=1\textwidth]{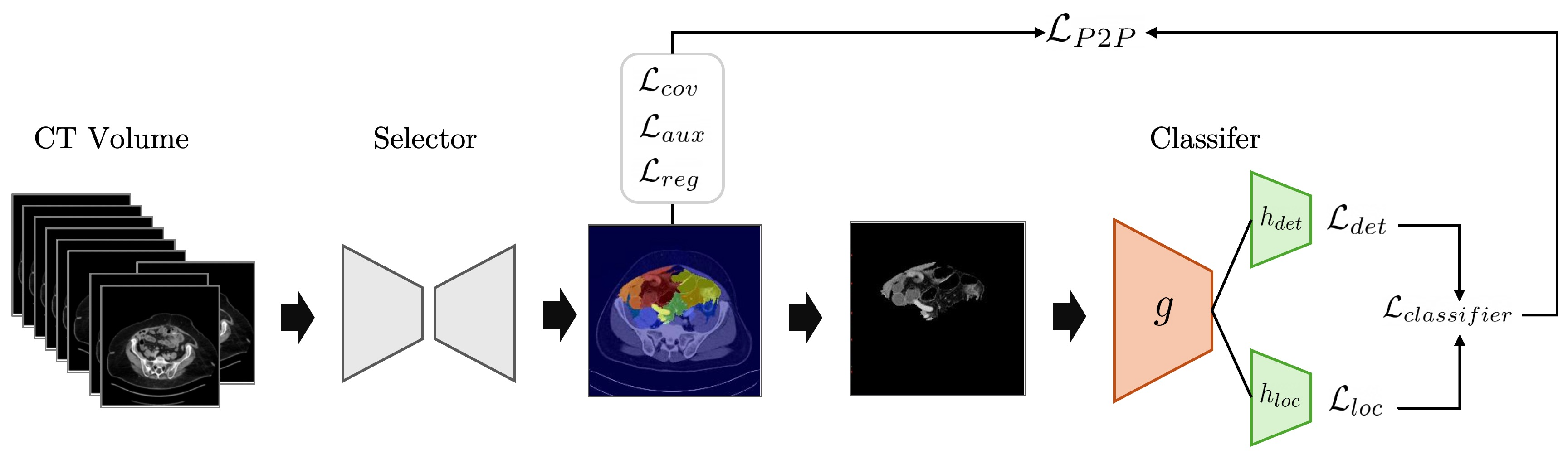}
    \caption{Schematic overview of the P2P extension. A slice is processed through the learnable selector that outputs a probabilistic selection mask. Subsequently, the masked input image is passed to the classifier as described in \Cref{fig:classifier}.}
    \label{fig:p2p}
\end{figure}
\paragraph{Training}
To operate at the slice level, each CT volume $\vx$ is separated into individual axial slices $\{\vx_s\}_{s=1}^S$. During training, each minibatch passed to the model $f$ comprises $B$ patients, where for each patient volume, a random slice is sampled. Each minibatch, hereafter denoted as $\vx$ for simplicity, is passed through a trainable backbone $g$ such as a ResNet~\citep{he2016deep} or a Vision Transformer~\citep{DosovitskiyZ21}, to obtain a latent representation $\vz=g(\vx)$. These features serve as a shared representation for the prediction of the different subtasks. As such, we utilize two prediction heads $h_{det}, h_{loc}$ that are responsible for detecting bowel obstruction and localizing the transition zone slices, respectively. Combined, $\{g, h_{det}, h_{loc}\}$ constitute the full model $f$ such that 
$$(\bm{\hat{y}}_{det}, \bm{\hat{y}}_{loc}) = f(\vx) = (h_{det}(\vz), h_{loc}(\vz)), \text{where } \vz=g(\vx).$$ 
In our work, we define $h$ as a simple linear probe consisting of a dropout layer followed by a linear layer. Notably, the dropout layer is not shared between $h_{det}$ and $h_{loc}$, thereby enriching the gradient signal for $\vz$. Finally, the training loss consists of two binary cross-entropy (BCE) terms $\mathcal{L}_{classifier} = \mathcal{L}_{det} + \lambda_{loc} \mathcal{L}_{loc}$. For obstruction detection, the patient-level label $y_{det} \in \{0,1\}$ is assigned to the corresponding slice, while for the transition zone localization, the slice-level labels $y_{loc} \in \{0,1\}$ are used directly. 
To decouple the two tasks, the localization loss is applied only to patients with obstruction. This strategy prevents $h_{loc}$ from implicitly performing obstruction detection and ensures that it is optimized specifically for transition zone localization.
Formally, the loss terms are defined as follows:
$$\mathcal{L}_{det} = \frac{1}{B}\sum_{i=1}^B\text{BCE}(y_{det}^{(i)}, \hat{y}_{det}^{(i)}), \quad \mathcal{L}_{loc} =  \frac{\sum_{i=1}^B\mathbbm{1}[y_{det}^{(i)} = 1] \text{BCE}(y_{loc}^{(i)}, \hat{y}_{loc}^{(i)})}{\sum_{i=1}^B \mathbbm{1}[y_{det}^{(i)} = 1]}$$

\paragraph{Inference}
Following the training of $f$, we now detail the inference procedure for patient-level detection and slice-level localization. To determine the presence of bowel obstruction in a patient, the full volume should be processed. As such, for a given patient, each slice $\{\vx_s\}_{s=1}^S$ is passed through $f$ to obtain predictions $\{\hat{y}_{det,s}\}_{s=1}^S$. Subsequently, the slice-level predictions are aggregated to a patient-level prediction $\hat{y}_{det} = \text{Agg}(\{\hat{y}_{det,s}\}_{s=1}^S)$. While more complex aggregation strategies based on $\vz$, such as post-training a bidirectional LSTM~\citep{hochreiter1997long,graves2005framewise}, can be considered, we find that in practice, a simple max aggregation suffices. Regarding the localization task, the model $h_{loc}$ generates a probability $p_s \in [0, 1]$ for each axial slice, representing the predicted likelihood of the transition zone's presence. In a clinical setting, a transition zone may span multiple contiguous slices. To align the model's output with this clinical reality, we treat the localization task as a retrieval problem of transition zone slices rather than a singular point prediction. Specifically, at inference time, all slices within the volume are ranked in descending order based on their predicted likelihoods:
$$\mathcal{R} = \text{sort\_desc}(\{\hat{y}_{loc,s}\}_{s=1}^S)$$
This ranked list $\mathcal{R}$ prioritizes the most suspicious slices, providing the clinician with an ordered sequence of candidates for the transition zone. By presenting results in this manner, the framework assists in rapid diagnostic verification by directing attention to the high-probability regions of the volumetric scan.

\subsection{Intra-Slice Localization}
\label{subsec:p2p_classifier}
This section describes an extension to the classifier $f$ based on an adaptation of the recently proposed P2P~\citep{vandenhirtz2025pixels} method. We provide a schematic overview in \Cref{fig:p2p}.
P2P is a method for training an inherently interpretable image classifier that faithfully visualizes the parts of the input relevant for its prediction. As such, the P2P framework is uniquely suited for faithfully localizing the transition point within a slice while retaining the dual-task capabilities of the classifier $f$. 
First, we provide the relevant background on P2P, then propose a series of domain-specific extensions that enable the application of this framework to a real-world medical dataset.

\paragraph{Preliminaries: P2P} 
This paragraph summarizes the work of \citet{vandenhirtz2025pixels}. For more details, we refer to their manuscript.
By enforcing a dependency on important regions, P2P ensures that a classifier's highlighted area is faithful to its prediction.
This dependency is enforced by masking out the unimportant parts of the image \textit{before} passing it to the classifier. This separates P2P from post-hoc explainability methods that attempt to recover the salient region \textit{after} a prediction was made~\citep{ribeiro2016should,selvaraju2017grad}. To achieve this, P2P introduces a probabilistic selector characterized by a learnable neural network. The selector operates on the level of superpixels~\citep{achanta2012slic} and predicts a binary selection mask $\vm$ for the image. This mask determines the regions in the input image that are masked out by $\vx_m=\vm \odot \vx$ before passing  $\vx_m$ to the classifier. Through the Gumbel-Softmax trick~\citep{Gumbel, concrete} this mask allows for gradients to flow from the classification task back to the selector. As such, the selector learns which regions are useful for prediction.
To prevent the selector from unmasking the entire image, a Kullback-Leibler (KL) divergence loss is utilized to regularize the average selection probability $\bar{p}$. At inference, a dynamic threshold determines the minimal area necessary for the model to reach a specific confidence level. Notably, P2P does not require segmentation annotations and learns the masking solely from the classification task.

\paragraph{P2P for Bowel Obstruction}
Transitioning P2P from curated benchmarks to real-world medical data necessitates several critical methodological changes, which we outline below.
First, we observe that the dynamic masking mechanism based on classification confidence is ill-suited for our task as a well-calibrated model should remain uncertain on a slice-level, therefore leading to the full image being displayed. Instead, we utilize a soft masking threshold $\tau$ that specifies the desired percentage of retained pixels. 
The second challenge is the fact that P2P was designed for multiclass problems. For a binary target, the risk of shortcut learning~\citep{geirhos2020shortcut} arises, where the selector bypasses region selection and trivially masks all pixels for negative cases.
We identify the root cause in the KL loss, which only regularizes the mean selection probability $\bar{p}$ for being too large, but not for being too small. Thus, the one-sided regularization loss is replaced with the Jensen-Shannon divergence $\mathcal{L}_{reg} = \mathcal{D}_{JS}(\bar{p} \parallel \tau)$ between $\bar{p}$ and the target threshold $\tau$, which penalizes the selector for both over- and under-masking. Finally, we leverage available transition point coordinate annotations to supervise the selector stage directly.
As P2P operates on a superpixel basis, we introduce an auxiliary intra-slice localization loss $\mathcal{L}_{aux}$ applied to the superpixel $j$ containing the transition point (TP):
$$\mathcal{L}_{aux}^{(i)} = \mathbbm{1}[TP^{(i)} \in \text{Superpixel}_j^{(i)} \wedge y_{det}^{(i)} = 1] \cdot \text{BCE}(1, p_j^{(i)}),$$
where $p_j$ is the selection probability for that patch. This ensures that the selector’s attention is grounded in diagnostic landmarks.

\bigskip\noindent
In summary, P2P serves as an enhancement to the classifier that preserves the model's detection and slice localization capabilities while also providing faithful intra-slice transition point detection. The final objective function of the P2P extension is defined as follows:
$$\mathcal{L}_{P2P} = \mathcal{L}_{classifier} + \lambda_2\mathcal{L}_{reg} + \lambda_3\mathcal{L}_{aux} + \lambda_4\mathcal{L}_{cov},$$
where $\mathcal{L}_{cov}$ denotes P2P's covariance loss that remains unchanged.
Ultimately, this framework enables an integrated clinical workflow where (\emph{i}) the model determines the presence of a bowel obstruction, (\emph{ii}) if positive, it identifies and ranks the most likely transition zone slices, and (\emph{iii}) it provides a faithful visualization of the suspected transition point region within those slices to assist in rapid diagnostic verification.

\section{Cohort}
This study follows a retrospective design and was approved by the local ethics committee. All patients provided informed consent for the use of their clinical data for research purposes.

\subsection{Cohort Selection}
\label{cohort_selection}

Consecutive adult patients ($\ge$ 18 years) with mechanical bowel obstruction identified on computed tomography (CT) were retrospectively included. Imaging studies were acquired between January 2014 and December 2024 at our institution. Both unenhanced and contrast-enhanced CT scans were considered. Mechanical bowel obstruction cases were defined based on radiological findings. Patients with paralytic ileus were excluded.  A control group of patients without evidence of bowel obstruction on CT imaging was included. Patients with malformed annotations were excluded.

\begin{figure}[t]
    \centering
    \begin{minipage}{0.47\linewidth}
        \centering
        \vspace{+1pt}
        \includegraphics[width=\linewidth]{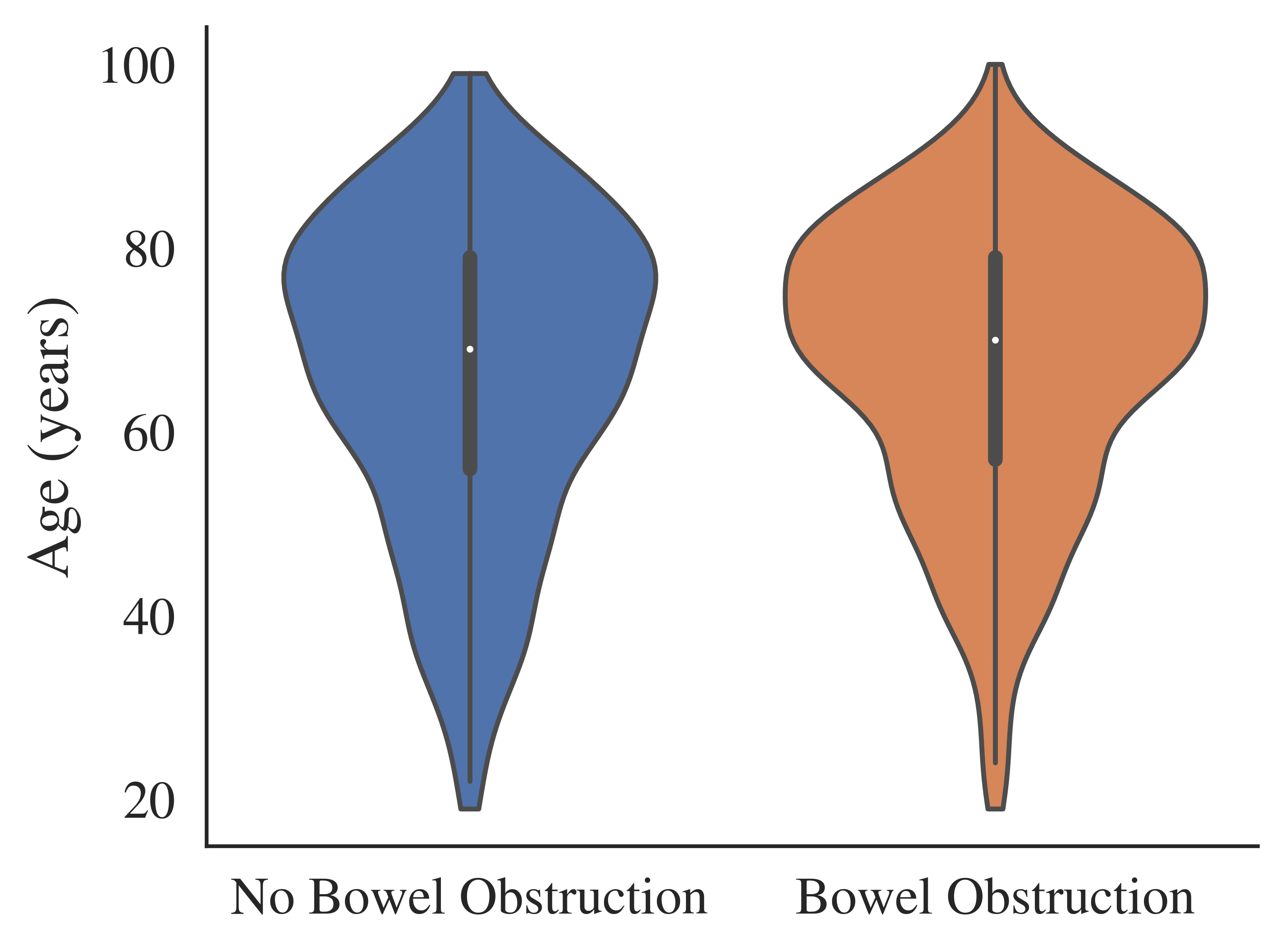}
        \caption*{(A) Age distribution}
    \end{minipage}
    \hfill
    \begin{minipage}{0.45\linewidth}
        \centering
        \vspace{-14pt}
        \includegraphics[width=\linewidth]{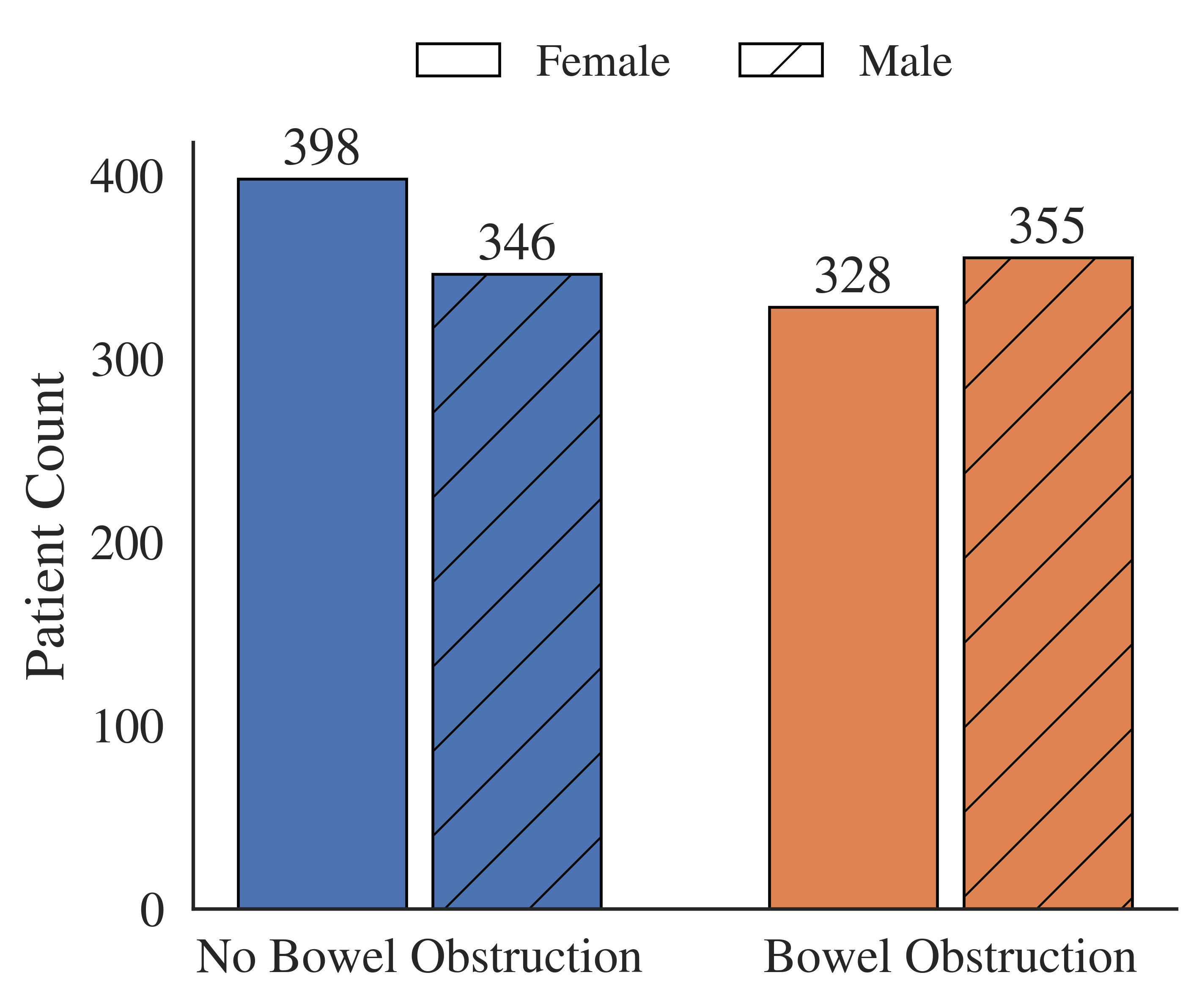}
        
        \caption*{(B) Sex distribution}
        
    \end{minipage}
    
    \caption{
     Cohort statistics. (A) Age distribution across patient groups. (B) Sex distribution across patient groups. The distributions indicate comparable age ranges and balanced sex composition between patients with and without bowel obstruction.
    }
    
    \label{fig:cohort}
\end{figure}

\bigskip\noindent
The final cohort comprised $1427$ patients, including $683$ patients with mechanical bowel obstruction and $744$ controls. The overall mean age was $67.0 \pm 16.4$ years. Patients with bowel obstruction had a mean age of $67.7 \pm 15.45$ years, compared to $66.4 \pm 17.0$ years in controls. Overall, $50.9\%$ of patients were female ($48.0\%$ in patients with mechanical bowel obstruction and $53.5\%$ in controls). Detailed age and sex distributions across patient groups are shown in Figure \ref{fig:cohort}. The data is not made public in compliance with privacy regulations.

\bigskip\noindent
The dataset was split at the patient level into training (80\%), validation (10\%), and test (10\%) sets. The training set was used for model development, while validation and test sets were used for model selection and final evaluation, respectively. A flowchart of the cohort selection process is shown in Appendix \ref{app:flowchart}.


\subsection{Data Extraction}

All imaging data were retrieved from the institutional Picture Archiving and Communication System (PACS). CT scans were acquired on multiple generations of scanners from a single vendor using standard clinical protocols, including portal venous or unenhanced acquisitions. Images were reconstructed in the axial plane with a cut thickness of 3 or 5 mm using a soft tissue kernel. 
Two board-certified radiologists reviewed all CT scans and identified the axial slice corresponding to the \textit{transition point} of the mechanical obstruction in each patient, when available. In the following chapters, we refer to this slice as the \textit{transition slice}. 
In this study, no additional clinical variables were used.

\subsection{Data Preprocessing}
\label{cohort_dataprocessing}

A \textit{transition zone} was defined as 10 consecutive slices centered on the radiologist-annotated transition slice. This set of slices better reflects clinical practice, as a mechanical obstruction is typically assessed over a short segment and may not be confined to a single slice. In Appendix \ref{app:slice_expansion}, we provide more details and analyze the sensitivity of our results to this choice.
Each axial slice was resized to $224 \times 224$ to balance computational efficiency and anatomical fidelity. Volumes and transition zone annotations were interpolated along the axial length dimension to a fixed depth of 60 slices, leading to an average of 4 transition zone slices per volume.
Intensity values were windowed using a mediastinal setting (center = 50 HU, width~=~350 HU; range $[-125, 225]$) and min--max normalized to $[0,1]$.

\section{Experimental Setup}
\label{experimental_setup}
This section describes the experimental protocol used to evaluate the proposed framework. We first define the evaluation tasks, followed by the baselines, evaluation metrics, and implementation details.

\subsection{Evaluation Tasks}

The proposed framework is evaluated along the three levels of its diagnostic hierarchy, corresponding to its intended clinical use. First, at the patient level, the model predicts whether a CT volume exhibits mechanical bowel obstruction, formulated as a binary classification task. Second, to assess performance on transition zone localization, we evaluate the model’s ability to rank and prioritize slices that are likely to contain the transition zone. This task measures whether the model effectively reduces the radiological search space by identifying a small subset of clinically relevant slices within the full CT volume. Third, at the intra-slice level, we assess the model’s ability to localize and highlight image subregions corresponding to the transition point. This task evaluates whether the model provides spatially meaningful and clinically interpretable evidence supporting its predictions.
Overall, this evaluation design reflects the radiological workflow, in which clinicians first diagnose the obstruction presence, then identify candidate slices and localize the precise transition point.

\subsection{Baselines}

To evaluate the proposed hierarchical framework, we compare it against baseline models that do not employ transition zone supervision. At the patient level, we train standard binary classification models using only patient-level labels indicating the presence or absence of bowel obstruction. As these models do not incorporate any transition-zone information, we use their slice-level obstruction detection probabilities to rank the slices by importance.
For intra-slice localization, we compare our P2P variant against explainability baselines and P2P without $\mathcal{L}_{aux}$ to isolate its effect.
We apply Grad-CAM~\citep{selvaraju2017grad} as a post-hoc method to our ResNet variant, as a comparison to $\text{P2P} - \mathcal{L}_{aux}$ as both do not use auxiliary transition point information. The Attention Rollout baseline~\citep{abnar2020quantifying} expands on our ViT-Ti variant, by training the attention-based heatmap with $\mathcal{L}_{aux}$. For region localization metrics requiring a hard threshold, the threshold is set such that on average, the same percentage of pixels $\bar{p}$ is activated as for P2P. 

\subsection{Evaluation Metrics}
\paragraph{Bowel obstruction detection}
Performance is evaluated using the area under the receiver operating characteristic curve (AUROC), which captures the model's ability to distinguish between patients with and without bowel obstruction across all classification thresholds. In addition, we report accuracy, sensitivity, and specificity at a fixed operating point for clinical interpretability. The operating threshold is selected on the validation set by maximizing Youden's index.

\paragraph{Transition Zone Localization}
To assess the models' ability to identify transition zone slices, the analysis centers on patients with bowel obstruction. Since the transition zone spans multiple slices and clinical utility depends on the relative ordering of candidate images, we frame this task as a retrieval problem. To this end, we utilize ranking-based metrics to evaluate how effectively the model localizes relevant transition-zone slices within the full CT volume. In ranking settings, Hit@10 is frequently used, as identifies the proportion of patients for which at least one positive slice is ranked within the top 10. Normalized Discounted Cumulative Gain (NDCG) measures the ranking quality by rewarding the presence of positive slices higher in the list. The score is normalized by the optimal ranking, providing a measure of how closely the model's output approaches a perfect retrieval.
As clinicians only need to locate a single slice within the transition zone to identify the full contiguous zone, we also report two metrics focused on the rank of the first correctly identified transition zone slice per patient. Mean Reciprocal Rank (MRR) calculates the average of the reciprocal rank of the best-ranked transition zone slice, which inherently discounts performance at later positions where the specific rank is of little consequence to a clinician's workflow. Lastly, MedRank captures the median position of the first correctly identified transition zone slice across the patients.

\paragraph{Intra-slice Transition Point Localization}
To evaluate transition point localization, we focus on the unique transition slice per patient for which annotated coordinates are available.
For a fair comparison across methods, we approximately match and report their mean percentage of active pixels $\bar{p}$. The first metric, Hit, measures whether the transition point (TP) was correctly localized within the highlighted region $\mathbbm{1}[TP \in Region]$, averaged over all positive patients. Additionally, we provide a weighted Hit (HitW) that weights correctly identified points by the slice's mask percentage $(1-p)\mathbbm{1}[TP \in Region]$. Finally, as P2P and the baselines all output a probabilistic selection mask, we report $\bar{p}_{min}$, representing the average minimum percentage of unmasked pixels required to include the transition point.

\subsection{Model Development Protocol}
To ensure fair comparison, all methods are trained and evaluated under the same data split, as defined in Section \ref{cohort_selection}. The preprocessing steps described in \Cref{cohort_dataprocessing} are applied consistently across all experiments. Hyperparameters are selected on the validation set. The test set evaluation is only performed once at the end of model development.

\subsection{Implementation Details}

All models are implemented in PyTorch~\citep{ansel2024pytorch}. We consider both convolutional and transformer-based architectures, namely a ResNet-50 backbone and Vision Transformers (ViTs) of sizes Tiny (ViT-Ti) and Base (ViT-B). All models are initialized with ImageNet-pretrained weights. 
During training, one axial slice per volume is sampled uniformly at random and augmented with random rotations up to $10^{\circ}$, applied with probability $0.8$. At evaluation time every slice of a volume is processed without rotation.
The model is trained jointly for patient-level detection and slice-level transition zone identification using a multi-task objective with a localization loss weight of $\lambda_{\text{loc}} = 1 \ \& \text{ lr}=2\times10^{-4}$ for ResNet and $\lambda_{\text{loc}} = 0.5 \ \& \text{ lr}=1\times10^{-4}$ for the ViT architectures. All models are trained with the Adam optimizer and a batch size of 64 for 100 epochs with model selection based on validation AUROC.

\begin{table*}[t]
\vspace{-0.05cm}
\centering
\caption{Performance (in \%) of baseline models and the proposed method for bowel obstruction detection and transition zone (TZ) localization.}
\label{tab:main}
\footnotesize
\setlength{\tabcolsep}{2.5pt}
\begin{tabular}{l l c c c c c c c c}
\toprule
\multirow{3}{*}{\textbf{Method}} & \multirow{3}{*}{\textbf{Backbone}} 
& \multicolumn{4}{c}{\textbf{Detection}} 
& \multicolumn{4}{c}{\textbf{TZ Localization}} \\
\cmidrule(lr){3-6} \cmidrule(lr){7-10}
& 
& AUROC $\uparrow$ & Acc. $\uparrow$ & Sens. $\uparrow$ & Spec. $\uparrow$ 
& Hit@10 $\uparrow$ & NDCG $\uparrow$ & MRR $\uparrow$ & MedRank $\downarrow$ \\
\midrule
\multirow{3}{*}{Baseline} & ViT-Ti & 95.35 & 89.43 & 83.33 & 93.90 & 73.33 & 50.95 & 31.95 & 6 \\
 & ViT-B & \textbf{96.69} & 88.73 & 81.67 & 93.90 & 75.00 & 51.97 & 32.51 & 6 \\
 & ResNet & 95.37 & 89.44 & 86.67 & 91.46 & 75.00 & 51.19 & 35.30 & 6 \\
\midrule
\multirow{3}{*}{Ours} & ViT-Ti & 94.74 & \textbf{92.96} & \textbf{90.00} & \textbf{95.12} & \textbf{95.00} & \textbf{62.32} & \textbf{48.51} & \textbf{3} \\
 & ViT-B & 94.94 & 89.44 & \textbf{90.00} & 89.02 & 85.00 & 61.08 & 46.59 & 3.5 \\
 & ResNet & 94.00 & 87.32 & 86.67 & 87.81 & \textbf{95.00} & 59.30 & 44.81 & 4 \\
\bottomrule
\vspace{-0.2cm}
\end{tabular}
\end{table*}

\section{Results and Discussion}

Following the experimental setup introduced in \Cref{experimental_setup}, the evaluation is structured along the three hierarchical axes of the proposed method: (i) patient-level obstruction detection, (ii) transition zone localization, and (iii) intra-slice transition point localization. 
As such, \Cref{tab:main} presents a comparative analysis of the proposed method against baseline models for obstruction detection and transition zone localization.
Then, \Cref{tab:p2p} presents the performance of the P2P extension, including intra-slice localization, and compares it to naively applying Grad-CAM to our method and to training with attention rollout.

\subsection{Obstruction Detection}

As shown in \Cref{tab:main}, all models exhibit strong obstruction detection performance. Baseline models, which were only trained on this task, achieve marginally higher overall AUROC. However, it is significant that the ViT-Tiny variant of our proposed dual-task framework demonstrates superior performance across all three threshold-dependent metrics (Accuracy, Sensitivity, and Specificity). 
This discrepancy between AUROC and classification metrics is noteworthy. While a higher AUROC indicates better general separability, the improved performance at the optimal Youden's index suggests that our method is more effective for real-world clinical application, where a definitive operating threshold is required. That is, in a diagnostic pipeline, the high sensitivity and specificity of the ViT-Tiny model may be preferred over the baselines, as it offers a more reliable classification. 
Furthermore, the P2P extension shown in \Cref{tab:p2p} exhibits an impressive detection performance of 96\% AUROC despite restricting the model's focus to a small portion of each slice. This indicates that P2P is able to identify highly predictive regions of the image, thereby enhancing trustworthiness and robustness without compromising diagnostic power.

\begin{figure}[t]
    \centering
    \hspace*{\fill} 
    \begin{minipage}[t]{0.48\textwidth}
        \centering
        \includegraphics[width=\textwidth]{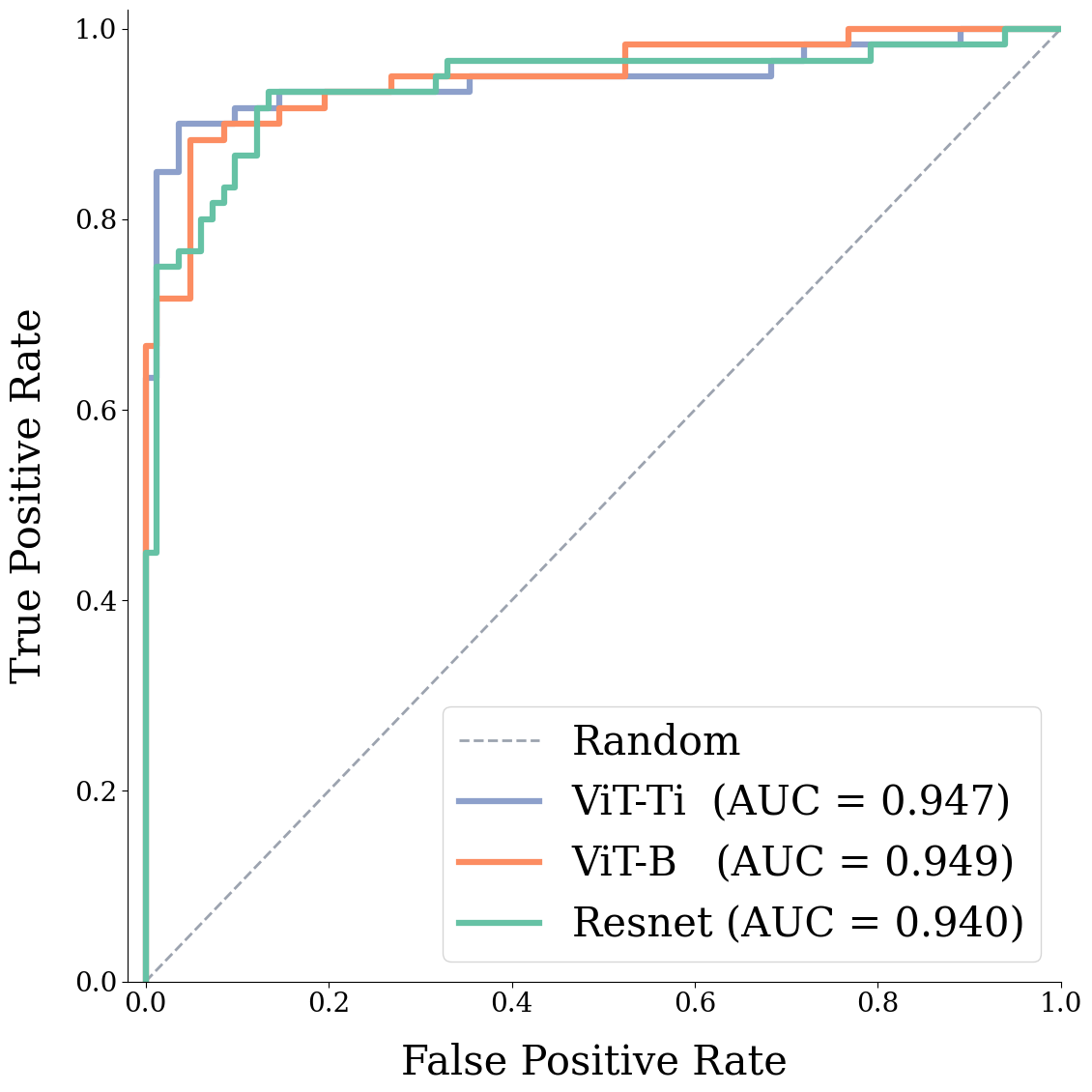}
        \caption{ROC curves on the test set for the proposed method across backbones.}
        \label{fig:auroc}
    \end{minipage}
    \hfill 
    \begin{minipage}[t]{0.48\textwidth}
        \centering
        \includegraphics[width=\textwidth]{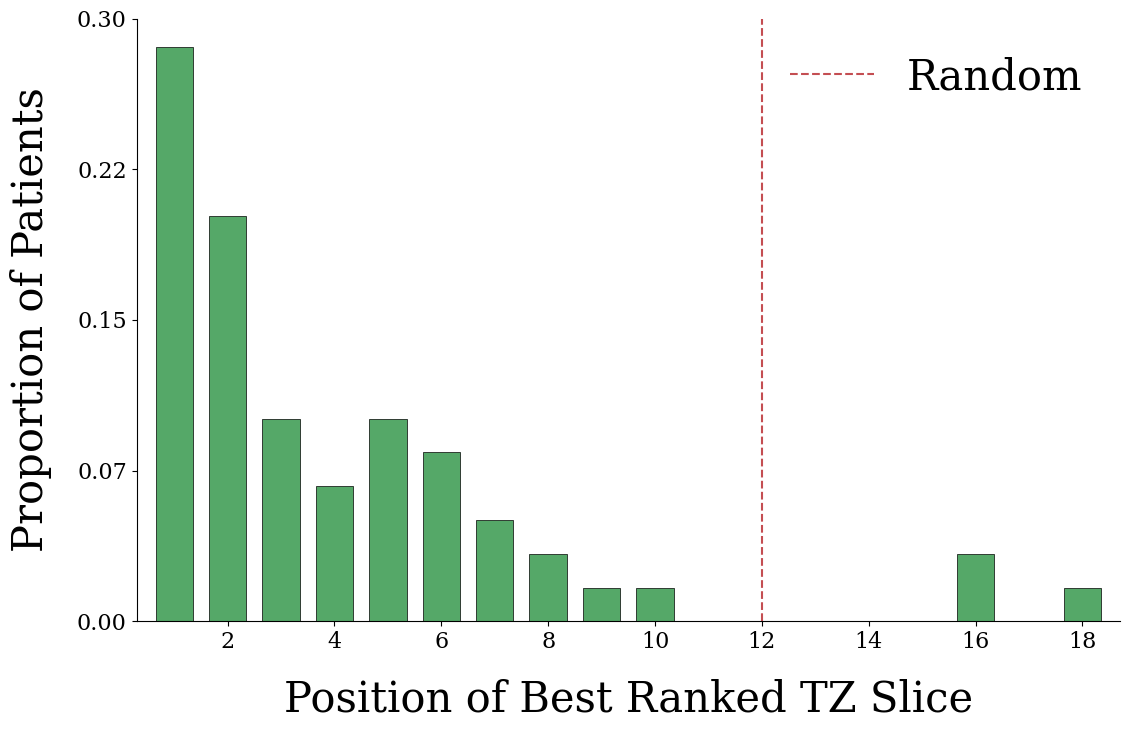}
        \caption{Rank distribution of the first correctly identified transition zone (TZ) slice per patient.}
        \label{fig:vit_performance}
    \end{minipage}
\end{figure}
\bigskip\noindent
In \Cref{fig:auroc}, we present the ROC curves on the test set for the three architectural variants of our method. Over all detection metrics, Vision Transformers (ViTs) appear better suited for our dual-task framework than the ResNet-50 backbone. This may be due to the global receptive field of the attention mechanism and the use of a distinct [CLS] token, which allows for a more flexible adaptation between global patient-level classification and local slice-level localization. Finally, existing literature on bowel obstruction detection with CTs presents test accuracies ranging from 73\%~\citep{oh2023deep,chang2026interpretable} to 86\%~\citep{vanderbecq2024deep} utilizing slice-wise and 3D CNN approaches. While their experimental results are not comparable as they differ in datasets, tasks, and methods, our test accuracy of 93\% is encouraging and demonstrates the potential efficacy of the proposed method.

\subsection{Transition Zone Localization}
\begin{figure}[ht]
    \centering
    \begin{minipage}{0.49\linewidth}
        \centering
        \includegraphics[width=\linewidth]{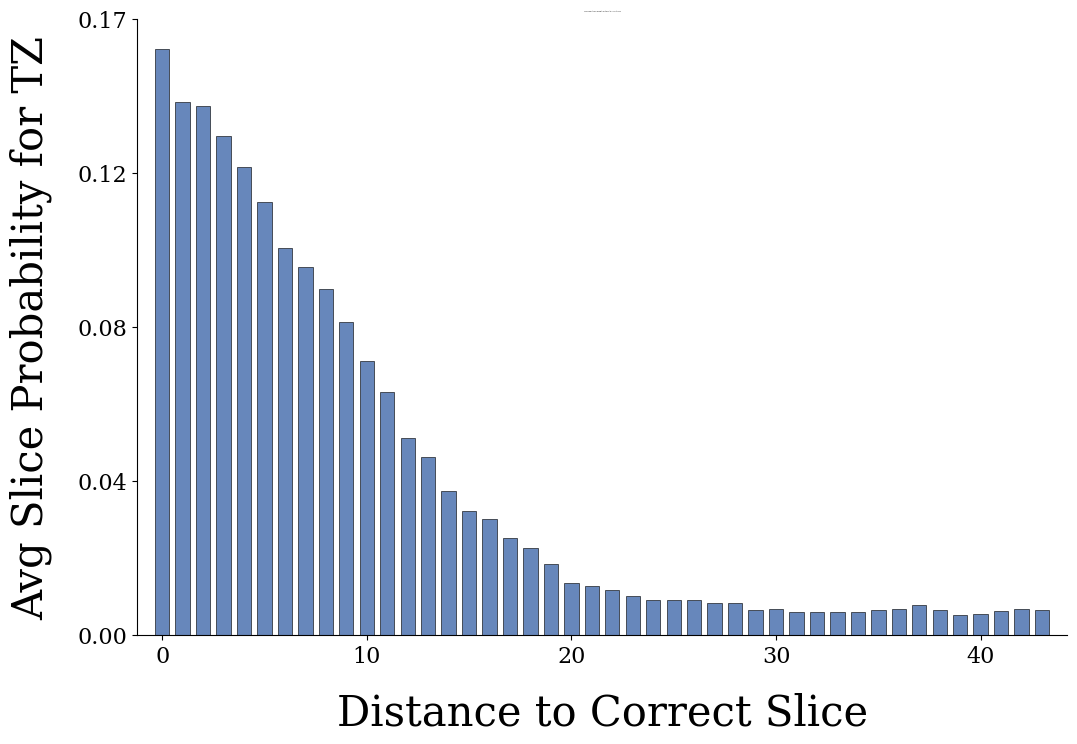}
        \caption*{(A) Ours (ViT-Tiny, TZ Head)}
    \end{minipage}
    \hfill
    \begin{minipage}{0.49\linewidth}
        \centering

        \includegraphics[width=\linewidth]{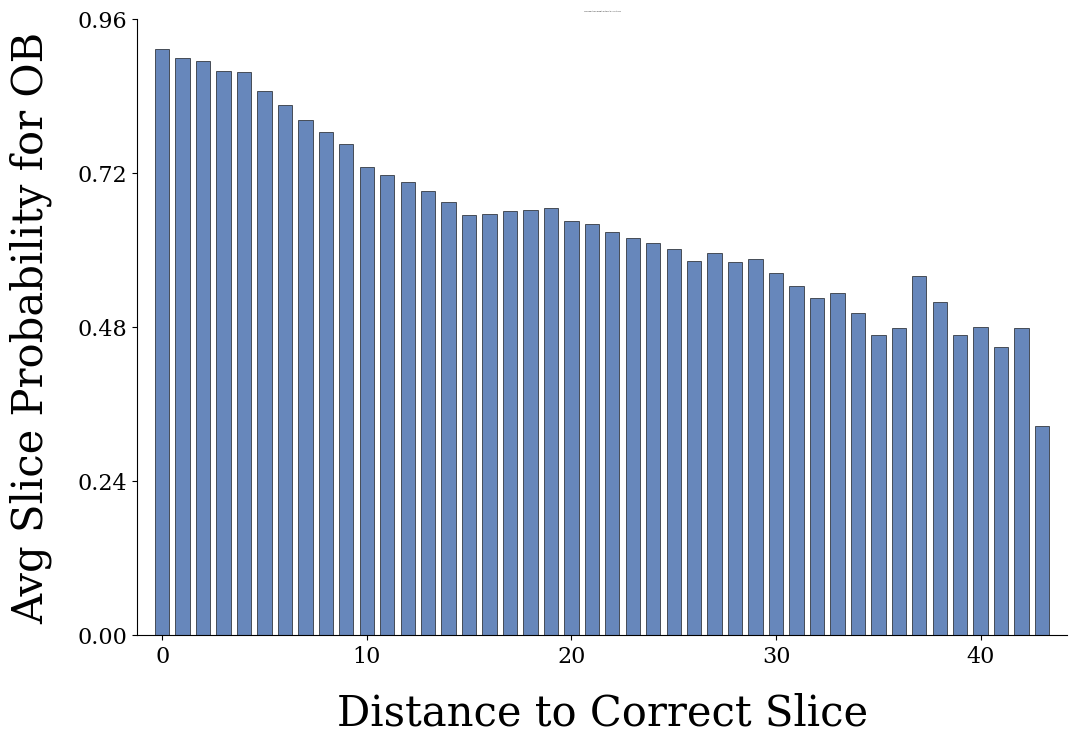}
        
        \caption*{(B) Baseline (ViT-Tiny, OB Head)}
        
    \end{minipage}
    
    \caption{Average slice probability as a function of distance to the nearest transition zone (TZ) slice. (A) Our ViT-Tiny model utilizing its transition zone (TZ) head. (B) The baseline ViT-Tiny model does not include a TZ head; therefore, we report slice-wise obstruction (OB) probabilities as a proxy.}
    
    \label{fig:sliceprobs}
\end{figure}




The transition zone localization in \Cref{tab:main} shows that the proposed method yields clear improvements. It consistently outperforms all baselines in Hit@10, NDCG, MRR, and MedRank across all backbones without compromising detection performance. The strongest results are obtained with ViT-Tiny (Hit@10 = 0.95, NDCG = 0.62, MRR = 0.49), together with the lowest median rank (MedRank = 3). These results indicate more accurate and reliable identification of transition zones. 
Remarkably in \Cref{tab:p2p}, despite being restricted to roughly 10\% of the image region, the P2P extension maintains competitive localization performance to its ViT-Tiny counterpart. To underscore the statistical reliability of \Cref{tab:main,tab:p2p}, we provide bootstrapped standard errors in Appendix~\ref{app:std}.
The clinical implications of these results are significant.
In practical terms, this means that instead of searching through a full CT volume, a radiologist only needs to examine a median of three slices to identify the transition zone. \Cref{fig:vit_performance} displays this rank distribution of the best-identified transition zone slices in more detail, showcasing that for nearly 30\% of patients, the highest ranked slice, contains the transition zone.
Furthermore, NDCG scores around 0.6 imply that the model's ranking of all transition slices substantially aligns with the optimal order. Finally, the Hit@10 score of 95\% indicates that for almost all cases, a transition zone slice can be found within the 10 highest-ranked slices. In Appendix~\ref{app:hitk}, we provide additional results for other Hit@K metrics.
Together, this suggests a strong reliability, demonstrating that the model's performance is not compromised by outlier cases.

\bigskip\noindent
In \Cref{fig:sliceprobs} we show that our ViT-Tiny model assigns high probabilities to slices near the transition zone, with a sharp decay as distance increases, indicating accurate and spatially coherent localization. In contrast, the baseline model's predictions are more uniformly distributed across slices, demonstrating that obstruction-based supervision alone is insufficient for accurate transition zone localization.
This observation paired with the fact that the baseline models perform worse on the localization metrics suggests that standard classifiers rely on secondary signs of obstruction rather than the transition point itself. 
By contrast, our joint-supervision framework successfully forces the model to learn the specific anatomy of the transition. Importantly, this does not come at the expense of detection performance as the proposed method's accuracy remains strong.
To our knowledge, this represents the first method to provide reliable transition zone localization, moving the technology closer to clinical feasibility.


\begin{table*}[t]
\vspace{-0.05cm}
\centering
\caption{
Performance (in \%) on bowel obstruction detection, transition zone (TZ) localization, and intra-slice transition point (TP) localization. Rows are grouped by $\mathcal{L}_{aux}$ usage. Grad-CAM and Attention Rollout use our ResNet and ViT-Ti variants, respectively.
}
\label{tab:p2p}
\footnotesize
\setlength{\tabcolsep}{2.5pt}
\begin{tabular}{l c c c c c c c c c c}
\toprule
\multirow{3}{*}{\textbf{Method}}  & \multirow{3}{*}{$\mathbf{\bar{p}}$}
& \multicolumn{2}{c}{\textbf{Detection}} 
& \multicolumn{4}{c}{\textbf{TZ Localization}} 
& \multicolumn{3}{c}{\textbf{TP Localization}}\\
\cmidrule(lr){3-4} \cmidrule(lr){5-8} \cmidrule(lr){9-11}
& 
& AUROC $\uparrow$ & Acc. $\uparrow$  
& Hit@10 $\uparrow$ & NDCG $\uparrow$ & MRR $\uparrow$ & MedRank $\downarrow$ & 
Hit $\uparrow$ & HitW $\uparrow$ & $\bar{p}_{min}$ $\downarrow$ \\
\midrule
Grad-CAM & 10.88 & 94.00 & 87.32 & \textbf{95.00} & 59.30 & 44.81 & 4 & 50.00 & 43.82 & 31.85 \\
P2P - $\mathcal{L}_{aux}$ &  12.59 & 94.21 & \textbf{91.55} & 90.00 & 60.82 & 45.26 & \textbf{3} & 68.33 & 59.60 & 9.27 \\
\midrule
Attn. Rollout & 10.52 & 95.57 & 90.85 & 86.67 & \textbf{61.31} & 44.53 & \textbf{3} & 83.33 & 74.33 & 5.55 \\
P2P &  10.48 & \textbf{96.30} & \textbf{91.55} & 86.67 & 61.01 & \textbf{49.20} & \textbf{3} & \textbf{88.33} & \textbf{79.06} & \textbf{4.84}  \\
\bottomrule
\vspace{-0.2cm}
\end{tabular}
\end{table*}
\begin{figure}
    \centering
    \includegraphics[width=1\textwidth]{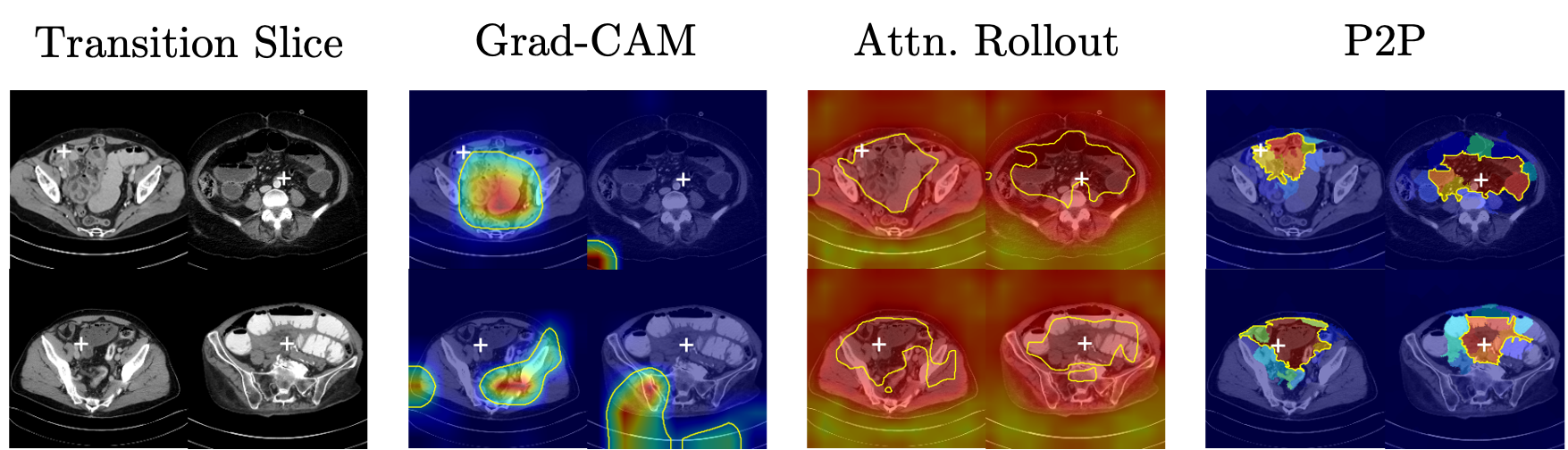}
    \caption{Qualitative comparison of intra-slice transition point localization on four selected examples ($2\times2$ grid). The transition point is marked with a white cross. Heatmaps are overlaid on the CT images. Yellow contours indicate each method's most salient regions.}
    \label{fig:p2p_qualitative}
\end{figure}

\subsection{Intra-Slice Transition Point Localization}

\looseness=-1
In \Cref{tab:p2p}, we present the intra-slice transition point localization performance of P2P and compare it to Grad-CAM's post-hoc explanations and Attention Rollout trained with $\mathcal{L}_{aux}$. P2P uses a ViT-Tiny backbone as it showed the strongest validation performance. Clearly, P2P excels at detecting the location of the transition point. This might be why it maintains strong detection and localization performance, as the focus on the transition point empowers the network to perform all tasks. Notably, a Hit metric of 88\% indicates that for most transition slices, P2P correctly localized the transition point while being constrained to just 10\% of the total image area on average. Even more significant is the fact that $\bar{p}_{min}$ is very low, indicating that the mean image area required to contain the transition point is just 5\%. 

\bigskip\noindent
In \Cref{fig:p2p_qualitative}, we show a qualitative comparison of the performance of P2P to Grad-CAM and Attention Rollout. Both Attention Rollout and P2P consistently highlight regions of the image that contain the transition point. However, the baselines produce rather diffuse activation maps, with multiple prominent regions and peak intensities (shown in red) that are either ambiguous or located away from the transition point. In contrast, P2P focuses more directly on the region containing the transition point, thereby yielding a spatially constrained and targeted importance map. In Appendix~\ref{app:qual_ex}, we provide additional examples that support our findings.

\bigskip\noindent
The superior performance of P2P in transition point localization is largely attributable to its inherently interpretable architecture, which permits the integration of the auxiliary transition point loss ($\mathcal{L}_{aux}$) during training. While the P2P variant without $\mathcal{L}_{aux}$ already generates better localization maps compared to the post-hoc Grad-CAM baseline, the inclusion of location-based supervision further optimizes the model's focus. 
Due to the benefit of $\mathcal{L}_{aux}$, Attention Rollout also performs well, albeit slightly worse than P2P in TP localization. Notably, due to P2P's inherent interpretability, the model's prediction is based on the highlighted region, while Attention Rollout provides no such guarantee. As such, P2P provides strong practical utility for failure mode analysis, model understanding, and trust.
While this necessitates the presence of transition point annotations, the P2P variant without auxiliary loss still shows promising localization. This provides evidence that our P2P extension by itself is already capable of identifying important regions that contain the transition point while never being explicitly trained for it.
By identifying the transition point within a minimal spatial region, the P2P extension positions itself as a reliable triage system that guides a radiologist’s attention to the most pathologically significant anatomy.


\section{Conclusion}
This work introduced a method to support clinicians in the automated diagnosis of bowel obstruction. The method performs patient-level obstruction detection as well as slice-level identification of the transition zone. Furthermore, an extension of P2P was proposed that enables the faithful localization of the transition point within a slice. Results on an internal dataset showed that the proposed method achieves a high diagnostic accuracy of 93\% while successfully localizing the transition zone within a median of just three slices. By leveraging a joint-supervision framework and an auxiliary loss, the P2P extension maintained strong detection while pinpointing the transition point within the slice using a minimal spatial region. 
In summary, by addressing the severely understudied task of transition zone localization, the proposed method marks a significant step towards automated identification of this important clinical landmark that would accelerate the diagnostic workflow and improve patient outcomes.

\paragraph{Limitations}
Despite the promising results, this study has several limitations. First, the evaluation was conducted on a retrospective, single-center dataset. As such, external validation on independent, multi-center cohorts is a necessary next step before the model's generalizability and clinical utility can be established; performance may vary across different patient populations, acquisition protocols, scanner vendors, and reconstruction settings, none of which were varied in the present study. Second, both bowel obstruction detection and transition zone localization are tasks characterized by high inter-observer variability. As such, our model remains dependent on a reference standard derived from subjective expert interpretations. Third, patients with paralytic ileus were not included in this study, limiting the representativeness of the clinical population. Future work should analyze such cases. Finally, while the proposed method and P2P extension proved effective for this specific gastrointestinal application, its generalizability to other complex medical problems and anatomical regions remains to be explored in future research.

\bibliography{sample}

\newpage
\appendix

\acks{MV is supported by the Swiss State Secretariat for Education, Research and Innovation (SERI) under contract number MB22.00047. AA is supported by the Swiss AI project ``Beyond Signals and Structure: Enabling Generalizable Multimodal Cardiopulmonary AI'' under contract number SNAI000046.}
\section{Hit@k Metrics}
\label{app:hitk}

Table~\ref{tab:app_hitk} reports Hit@1, Hit@5, and Hit@10 for all baselines and proposed model variants.
Across all backbones, the proposed method substantially improves the model's ability to rank the true transition slice at or near the top: Hit@1 nearly doubles relative to the baseline, and this gap persists at Hit@5. This indicates that the improvement over the baseline is not solely driven by broader retrieval (Hit@10) but also by more precise top-rank localization.

\begin{table}[h]
\centering
\caption{TZ localization performance at stricter rank thresholds (in \%).}
\label{tab:app_hitk}
\small
\begin{tabular}{l l c c c}
\toprule
\textbf{Method} & \textbf{Backbone} & \textbf{Hit@1 $\uparrow$} & \textbf{Hit@5 $\uparrow$} & \textbf{Hit@10 $\uparrow$} \\
\midrule
\multirow{3}{*}{Baseline} & ViT-Ti  & 15.00 & 48.33 & 73.33 \\
                          & ViT-B   & 15.00 & 48.33 & 75.00 \\
                          & ResNet  & 20.00 & 48.33 & 75.00 \\
\midrule
\multirow{3}{*}{Ours}     & ViT-Ti  & 28.33 & \textbf{75.00} & \textbf{95.00} \\
                          & ViT-B   & 31.67 & 65.00 & 85.00 \\
                          & ResNet  & 30.00 & 63.00 & \textbf{95.00} \\
\midrule
P2P (Ours)                 & ViT-Ti  & \textbf{33.33} & 71.67 & 86.67 \\
\bottomrule
\end{tabular}
\end{table}

\section{Sensitivity to Transition-Zone Label Width}
\label{app:slice_expansion}

Because the dataset provides only a point annotation for the transition zone, we construct positive labels by expanding a fixed number of slices around the annotated point, reflecting our best clinical estimate of the true extent of the transition zone. Afterward, the CT volume is resized to $60$ axial slices via bilinear interpolation. The corresponding slice-level labels, however, cannot be interpolated directly, since they are binary; instead, we map them via nearest-neighbor index rescaling: each original slice index is rescaled to the resized volume's index range, and a resized slice is marked positive if it is the nearest match to any originally positive slice. To assess how sensitive our reported results are to this choice, Table~\ref{tab:app_expansion} reports Hit@10, NDCG, MRR, and median rank for slice expansions of 2, 5, and 10 slices, along with the resulting average number of positive slices per study after resampling.

\bigskip\noindent
Performance improves monotonically as the label width increases, which is expected: a wider positive region is an easier retrieval target under rank-based metrics. Importantly, even under the narrowest, most conservative label width (expansion of 2, yielding on average only 2.25 positive slices post-resampling), the model still achieves a Hit@10 of 76.67\% and a median rank of 5, confirming that our main results are not an artifact of a generously wide label definition. We determined an expansion of 5 as the default independently of model development, as it best matches our estimate of the physical extent of a transition zone, balancing label fidelity against overly narrow point-level supervision.

\begin{table}[h]
\centering
\caption{Sensitivity of TZ localization performance to slice-expansion width used for label construction (in \%, except MedianRank and Avg.\ Correct Slices).}
\label{tab:app_expansion}
\small
\begin{tabular}{l c c c c c}
\toprule
\textbf{Slice Expansion} & \textbf{Hit@10 $\uparrow$} & \textbf{NDCG $\uparrow$} & \textbf{MRR $\uparrow$} & \textbf{MedRank $\downarrow$} & \textbf{Avg.\ Correct Slices} \\
\midrule
2            & 76.67 & 49.25 & 31.99 & 5 & 2.25 \\
5 (Default)  & 95.00 & 62.32 & 48.51 & 3 & 4.15 \\
10           & 96.67 & 69.29 & 54.92 & 2 & 6.95 \\
\bottomrule
\end{tabular}
\end{table}

\section{Bootstrapped Performance Estimates}
\label{app:std}
To quantify the uncertainty of the performance estimators due to the test set's finite sample size, we estimate the variability of our results under resampling of the test set. 
Specifically, we generate $10{,}000$ bootstrap samples by resampling patients from the test set with replacement, and recompute all metrics on each resample. For every metric, this yields both a point estimate and its standard error, the latter quantifying the uncertainty in how well our test-set estimate reflects the performance on the broader underlying patient distribution.

\bigskip\noindent
Tables~\ref{tab:app_std_main} and~\ref{tab:app_std_p2p} report the bootstrapped point estimates and standard errors corresponding to Tables~\ref{tab:main} and~\ref{tab:p2p} in the main text, respectively. We note that the point estimates for MedRank differ slightly from those reported in the main text; this is an expected consequence of bootstrap resampling due to the non-continuous nature of the median-based metric and does not affect any of our conclusions. Across both tables, the standard errors are consistently small relative to the differences between methods, indicating that the reported performance gains are unlikely to be an artifact of the particular composition of our test set, though naturally subject to some residual uncertainty.

\begin{landscape}
\begin{table}
\centering
\caption{Performance (in \%) of baseline models and the proposed method for bowel obstruction detection and transition zone (TZ) localization, with bootstrapped standard errors.}
\label{tab:app_std_main}
\scriptsize
\setlength{\tabcolsep}{3pt}
\begin{tabular}{l l c c c c c c c c}
\toprule
\multirow{3}{*}{\textbf{Method}} & \multirow{3}{*}{\textbf{Backbone}} 
& \multicolumn{4}{c}{\textbf{Detection}} 
& \multicolumn{4}{c}{\textbf{TZ Localization}} \\
\cmidrule(lr){3-6} \cmidrule(lr){7-10}
& 
& AUROC $\uparrow$ & Acc. $\uparrow$ & Sens. $\uparrow$ & Spec. $\uparrow$ 
& Hit@10 $\uparrow$ & NDCG $\uparrow$ & MRR $\uparrow$ & MedRank $\downarrow$ \\
\midrule
\multirow{3}{*}{Baseline} 
& ViT-Ti & 95.35 $\pm$ 1.98 & 89.43 $\pm$ 2.56 & 83.33 $\pm$ 4.87 & 93.90 $\pm$ 2.64 & 73.33 $\pm$ 5.71 & 50.95 $\pm$ 2.26 & 31.95 $\pm$ 4.11 & 5.96 $\pm$ 1.10 \\
& ViT-B  & \textbf{96.69} $\pm$ 1.28 & 88.73 $\pm$ 2.62 & 81.67 $\pm$ 5.06 & 93.90 $\pm$ 2.61 & 75.00 $\pm$ 5.64 & 51.97 $\pm$ 2.31 & 32.51 $\pm$ 4.07 & 5.60 $\pm$ 1.07 \\
& ResNet & 95.37 $\pm$ 1.96 & 89.44 $\pm$ 2.57 & 86.67 $\pm$ 4.43 & 91.46 $\pm$ 3.09 & 75.00 $\pm$ 5.59 & 51.19 $\pm$ 2.34 & 35.30 $\pm$ 4.48 & 5.73 $\pm$ 0.92 \\
\midrule
\multirow{3}{*}{Ours} 
& ViT-Ti & 94.74 $\pm$ 2.27 & \textbf{92.96} $\pm$ 2.12 & \textbf{90.00} $\pm$ 3.90 & \textbf{95.12} $\pm$ 2.37 & \textbf{95.00} $\pm$ 2.79 & \textbf{62.32} $\pm$ 2.37 & \textbf{48.51} $\pm$ 4.51 & \textbf{2.71} $\pm$ 0.63 \\
& ViT-B  & 94.94 $\pm$ 1.93 & 89.44 $\pm$ 2.56 & \textbf{90.00} $\pm$ 3.90 & 89.02 $\pm$ 3.46 & 85.00 $\pm$ 4.57 & 61.08 $\pm$ 2.68 & 46.59 $\pm$ 4.92 & 3.51 $\pm$ 0.81 \\
& ResNet & 94.00 $\pm$ 2.28 & 87.32 $\pm$ 2.75 & 86.67 $\pm$ 4.41 & 87.81 $\pm$ 3.58 & \textbf{95.00} $\pm$ 2.79 & 59.30 $\pm$ 2.46 & 44.81 $\pm$ 4.92 & 4.13 $\pm$ 0.63 \\
\bottomrule
\end{tabular}
\vspace{2cm}
\caption{Performance (in \%) on bowel obstruction detection, transition zone (TZ) localization, and intra-slice transition point (TP) localization, with bootstrapped standard errors. Rows are grouped by $\mathcal{L}_{aux}$ usage.}
\label{tab:app_std_p2p}
\begin{tabular}{l c c c c c c c c c c c c}
\toprule
\multirow{2}{*}{\textbf{Method}}  & \multirow{2}{*}{\textbf{Backbone}} & \multirow{2}{*}{$\mathbf{\bar{p}}$}
& \multicolumn{2}{c}{\textbf{Detection}} 
& \multicolumn{4}{c}{\textbf{TZ Localization}} 
& \multicolumn{3}{c}{\textbf{TP Localization}}\\
\cmidrule(lr){4-5} \cmidrule(lr){6-9} \cmidrule(lr){10-12}
& & 
& AUROC $\uparrow$ & Acc. $\uparrow$  
& Hit@10 $\uparrow$ & NDCG $\uparrow$ & MRR $\uparrow$ & MedRank $\downarrow$ & 
Hit $\uparrow$ & HitW $\uparrow$ & $\bar{p}_{min}$ $\downarrow$ \\
\midrule
Grad-CAM     & ResNet & 10.88 & 94.00 $\pm$ 2.28 & 87.32 $\pm$ 2.75 & \textbf{95.00} $\pm$ 2.79 & 59.30 $\pm$ 2.46 & 44.81 $\pm$ 4.92 & 4.13 $\pm$ 0.63 & 50.00 $\pm$ 6.44 & 43.82 $\pm$ 5.66 & 31.85 $\pm$ 4.93 \\
P2P - $\mathcal{L}_{aux}$ & ResNet & 12.59 & 94.21 $\pm$ 2.30 & \textbf{91.55} $\pm$ 2.41 & 90.00 $\pm$ 3.90 & 60.82 $\pm$ 2.55 & 45.26 $\pm$ 4.57 & 3.22 $\pm$ 0.61 & 68.33 $\pm$ 6.01 & 59.60 $\pm$ 5.24 & 9.27 $\pm$ 0.87 \\
\midrule
Attn.\ Rollout & ViT-Ti & 10.52 & 95.57 $\pm$ 1.96 & 90.85 $\pm$ 2.40 & 86.67 $\pm$ 4.37 & \textbf{61.31} $\pm$ 2.48 & 44.53 $\pm$ 4.49 & 3.27 $\pm$ 0.71 & 83.33 $\pm$ 4.80 & 74.33 $\pm$ 4.30 & 5.55 $\pm$ 0.61 \\
P2P          & ViT-Ti & 10.48 & \textbf{96.30} $\pm$ 1.51 & \textbf{91.55} $\pm$ 2.31 & 86.67 $\pm$ 4.42 & 61.01 $\pm$ 2.56 & \textbf{49.20} $\pm$ 4.92 & \textbf{3.21} $\pm$ 0.80 & \textbf{88.33} $\pm$ 4.22 & \textbf{79.06} $\pm$ 3.79 & \textbf{4.84} $\pm$ 0.57 \\
\bottomrule
\end{tabular}
\end{table}
\end{landscape}

\section{Additional Qualitative Examples}
\label{app:qual_ex}
\begin{figure}[h]
    \centering
    \includegraphics[width=0.5\textwidth]{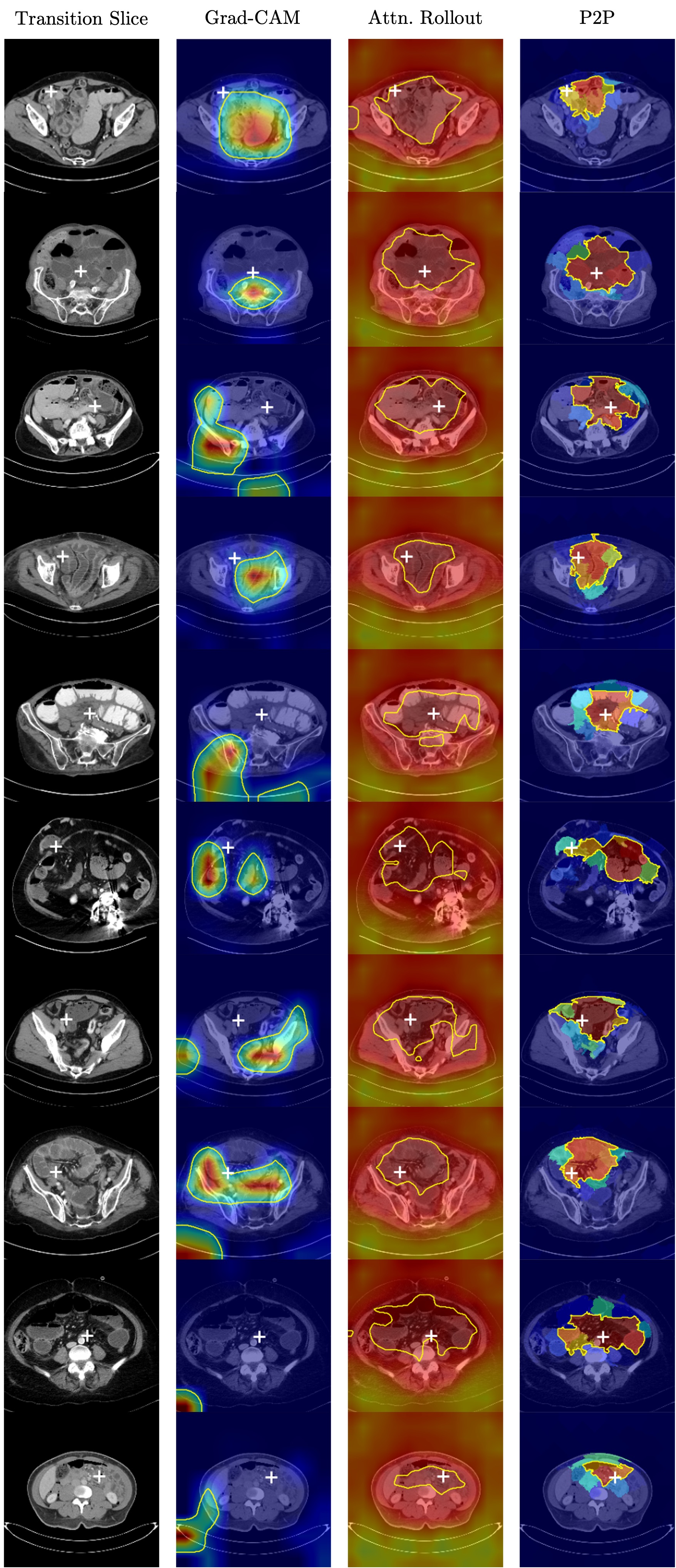}
    \caption{Qualitative comparison of intra-slice transition point localization on ten selected examples. The transition point is marked with a white cross. Heatmaps are overlaid on the CT images. Yellow contours indicate each method's most salient regions.}
    \label{fig:p2p_qualitative_app}
\end{figure}

In \Cref{fig:p2p_qualitative_app}, we present additional qualitative examples for intra-slice transition point localization of all methods.

\section{Cohort Selection}
\label{app:flowchart}
\begin{figure}[h]
   \centering 
   \includegraphics[width=4.5in]{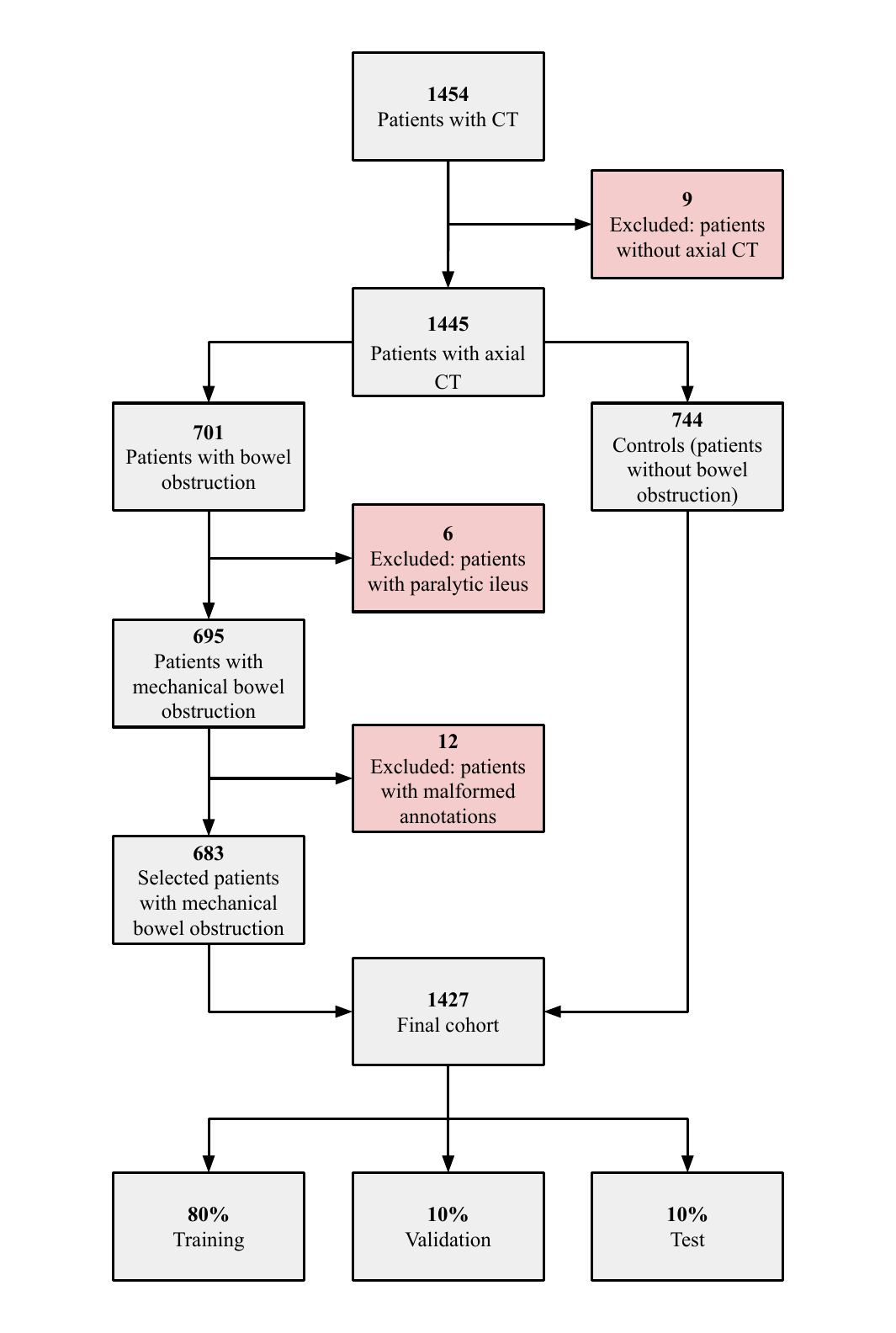} 
   \caption{\textbf{Cohort selection flowchart}. After excluding non-axial scans, patients were stratified into obstruction and control groups. Paralytic bowel and cases with invalid annotations were removed, yielding a final cohort of 1427 patients, split into training, validation, and test sets (80\%/10\%/10\%).}
   \label{fig:flowchart} 
 \end{figure}

In \Cref{fig:flowchart}, we present a flowchart of the cohort selection process.
\end{document}

%% file: math_commands.tex
\usepackage{amsmath,amsfonts,bm}

\def\eqref#1{equation~\ref{#1}}

\def\1{\bm{1}}

\def\vm{{\bm{m}}}

\def\vx{{\bm{x}}}

\def\vz{{\bm{z}}}

\DeclareMathAlphabet{\mathsfit}{\encodingdefault}{\sfdefault}{m}{sl}
\SetMathAlphabet{\mathsfit}{bold}{\encodingdefault}{\sfdefault}{bx}{n}

